\documentclass[10pt,twocolumn,letterpaper]{article}
\usepackage{wacv}
\usepackage{times}
\usepackage{epsfig}
\usepackage{graphicx}
\usepackage{amsmath}
\usepackage{amssymb}
\usepackage{booktabs}
\usepackage{subcaption}
\usepackage{authblk}

\wacvfinalcopy 

\ifwacvfinal
\usepackage[breaklinks=true,bookmarks=false]{hyperref}
\else
\usepackage[pagebackref=true,breaklinks=true,colorlinks,bookmarks=false]{hyperref}
\fi

 \title{Deep Metric Learning with Soft Orthogonal Proxies}

\author{Farshad Saberi-Movahed$^{1}$\hspace{0.1mm}, Mohammad K.Ebrahimpour$^{2}$\hspace{0.1mm}, Farid Saberi-Movahed$^{3}$\hspace{0.1mm}, Monireh Moshavash$^{4}$\hspace{0.1mm}, Dorsa Rahmatian$^{4}$\hspace{0.1mm}, Mahvash Mohazzebi$^{4}$\hspace{0.1mm}, Mahdi Shariatzadeh$^{4}$\hspace{0.1mm}, Mahdi Eftekhari$^{4}$ \\
$^{1}$NVIDIA, Santa Clara, CA 95051, USA,
$^{2}$Ericsson,Santa Clara,CA 95054,USA, 
$^{3}$Department of Applied Mathematics, Graduate University of Advanced Technology, Kerman, Iran, 
$^{4}$Department of Computer Engineering, Shahid Bahonar University of Kerman, Kerman, Iran 

{\tt\small fmovahed@nvidia.com, mkebrahimpour@gmail.com}
}
 
\begin{document}
\maketitle
\thispagestyle{empty}

\begin{abstract}
Deep Metric Learning (DML) models rely on strong representations and similarity-based measures with specific loss functions. Proxy-based losses have shown great performance compared to pair-based losses in terms of convergence speed. However, proxies that are assigned to different classes may end up being closely located in the embedding space and hence having a hard time to distinguish between positive and negative items. Alternatively, they may become highly correlated and hence provide redundant information with the model. To address these issues, we propose a novel approach that introduces Soft Orthogonality (SO) constraint on proxies. The constraint ensures the proxies to be as orthogonal as possible and hence control their positions in the embedding space. Our approach leverages Data-Efficient Image Transformer (DeiT) as an encoder to extract contextual features from images along with a DML objective. The objective is made of the Proxy Anchor loss along with the SO regularization. We evaluate our method on four public benchmarks for category-level image retrieval and demonstrate its effectiveness with comprehensive experimental results and ablation studies. Our evaluations demonstrate the superiority of our proposed approach over state-of-the-art methods by a significant margin.
\end{abstract}

\section{Introduction}
Deep Metric Learning (DML)~\cite{kaya2019deep} is a widely employed algorithm in similarity-based computer vision tasks such as image retrieval~\cite{cakir2019deep,passalis2020variance}, image clustering~\cite{sohn2016improved,wang2019ranked}, and person re-identification~\cite{yi2014deep,cheng2016person,wang2018equidistance}. DMLs utilize deep neural networks (DNNs) as an encoder to transform images into an embedding space that is representative of images and reflects their similarities accurately using a specific distance metric. The embedding space is usually learned through a supervised learning technique using an appropriate loss function.

In the past few years, the majority of proposed DML methods in the literature have leveraged Convolutional Neural Networks (CNNs) as the encoder network~\cite{ebrahimpour_ijcnn,ebrahimpour_wacv,ebrahimpour_dml}. However, a few works have recently investigated the performance of transformers~\cite{el2021training} on encoding images into a proper embedding space. In particular, El-Nouby \etal~\cite{el2021training} applied a particular Vision Transformer (ViT)~\cite{dosovitskiy2020image} called Data-Efficient Image Transformer (DeiT) ~\cite{touvron2021training} to the image retrieval task. ViTs are novel DNNs that adopt the transformer~\cite{vaswani2017attention} concepts in the Natural Language Processing (NLP) for computer vision tasks. Numerous studies have shown the impressive performance of transformers in NLP and Natural Language Understanding (NLU) tasks which have led to an explosion in generative artificial intelligence (AI) applications. As opposed to CNNs, ViTs do not have any typical convolutional layers and hence can result in different inductive biases and distinct learning compared to CNNs. More specifically, the self-attention layers in ViTs can attend to different parts of the input data and extract contextualized and semantic information from the input data~\cite{el2021training}.

In addition to the DNN architecture, the loss function used to train the encoder is another important part of any DML model. The loss functions are essential to provide a necessary supervisory signal based on the problem objectives~\cite{kim2020proxy,wang2019multi}. Recently, proxy-based loss functions have led to DML models with impressive results. The proxy-based losses relax the original problem and introduce proxies for similarity comparisons rather than comparing the pairwise similarities between data points. Therefore, they converge faster than pair-based models. However, one potential caveat with proxy-based models is that the proxies associated with different classes end up being close to each other in the embedding space or in other cases some proxies are redundant. Redundancy within proxy vectors is when some of the proxies are highly correlated with each other, providing repetitive information to the model. This can be problematic because it increases computational complexity and reduces the effectiveness of the training process, which makes the training process less effective by guiding the model to focus on obtaining similar proxies instead of learning more diverse representations. As a result, the proxy-based models can be negatively affected by the redundancy in proxy vectors. 

There are some successful ways that are introduced in the literature to recognize and remove redundant proxies. One strategy is to combine regularization terms with the proxy-based loss function to facilitate the sparseness and separation of proxies in the embedding space. The Soft Orthogonality (SO)~\cite{bansal2018can} is a simple regularization technique that can potentially drive a proxy-based DML algorithm to learn proxy vectors to be as orthogonal as possible so that they can distinguish between similar yet distinct examples. Thus, it inspired us to explore the effects of integrating the SO regularization approach with the proxy-based losses.

In this paper, we propose a constrained Proxy-based Image Retrieval Transformer (PIRT) to overcome the limitations of the proxy-based models. Our proposed method adopts DeiT as the backbone of the encoder to map the raw images to a latent space. Then, a constrained proxy-based loss is proposed to ensure the proxies in the latent space are as orthogonal to each other as possible. This is achieved by adding a regularization term based on the SO constraint to the loss function. This regularization term, grounded on the Frobenius norm of the Gram matrix of proxy vectors, determines the total squared magnitude of the dot products among all pairs of proxy vectors. The SO regularization in the PIRT loss function leads to attaining highly orthogonal proxy vectors and eliminating redundancy in them as much as possible. This makes learned proxy vectors be more discriminative and less likely to confuse similar classes. We thoroughly evaluated our proposed PIRT method on four publicly available benchmarks.

The rest of the paper is organized as follows. Section \ref{sec:2} reviews the related works. Section \ref{sec:3} presents our proposed algorithm to use a constraint Proxy Anchor loss to train an embedding layer attached on top of a vision transformer for image retrieval. Section \ref{Experimental Results} demonstrates the experimental results where the benchmark datasets are introduced, and quantitative and qualitative results are discussed. Section \ref{concsec} concludes the presented work with some suggestions for the future work.

\section{Related Work}\label{sec:2}
\textbf{DML Loss Functions:}
The loss functions used in DML can be mainly grouped into two categories, pair-based, and proxy-based loss functions. The pair-based loss functions use pairwise distances between data points in the embedding space to derive signals for training. The triplet loss~\cite{wang2014learning,hoffer2015deep,schroff2015facenet}, contrastive loss~\cite{bromley1993signature,chopra2005learning,hadsell2006dimensionality}, and Multi Similarity (MS) loss~\cite{wang2019multi} are the well-known pair-based loss functions. Despite the rich supervisory signals obtained from the data-to-data comparisons in these loss functions, they suffer from a high computational overhead that leads to slow convergence. Furthermore, due to random sampling, the training data often contain trivial, less informative pairs that can exacerbate the slow convergence issue and even model degeneration. Although some sampling techniques such as hard negative mining can resolve the trivial samples problem to some extent, it still requires some manual hyperparameter tuning for tuple sampling.

The proxy-based losses have been introduced to solve the slow convergence issue of the pair-based losses. Their main idea is to consider a proxy for each subset of data (e.g., one proxy for each class of data) in the form of a vector, which is learned during the training process, with the same shape as the embedding vectors. Movshovitz-Attias \etal~\cite{movshovitz2017no} introduced Proxy-NCA that treats each data point as an anchor and associates it with the proxies rather than other data points. Then, the loss function tries to bring the anchors closer to the proxy of the same class and drives away those from the proxies of different classes in the embedding space. Because the number of proxies is significantly smaller than that of data points, the training complexity of Proxy-NCA is improved compared to pair-based losses. However, the rich data-to-data relation signals of the pair-based losses are lost in the Proxy-NCA loss due to relying on the data-to-proxy relations.

To alleviate the Proxy-NCA’s issue described above, the Proxy Anchor loss~\cite{kim2020proxy} has been proposed to leverage the best of both worlds. In the Proxy Anchor loss, each proxy is an anchor which is associated with all the data in the training batch. To capture the data-to-data relation, the gradient of the loss with respect to a data point is weighted by its relative proximity to the proxy which is affected by the other data points in the batch. The farther the proxy and the positive example are, the more strongly they pull together, and the closer the proxy and negative example are, the more strongly they push away.

\textbf{Vision Transformers:}
In a groundbreaking work, Vaswani et al.~\cite{vaswani2017attention} proposed the transformer network that incorporated the self-attention mechanism to overcome the issues of the Long-Short Term Memory (LSTM) networks~\cite{hochreiter1997long,sutskever2014sequence,sherstinsky2020fundamentals} such as the memory loss on long sequences in machine translation applications~\cite{hernandez2021attention}. It soon became the basis of many innovations in the fields of NLP and NLU, which gave birth to popular models such as Bidirectional Encoder Representations from Transformers (BERT) models~\cite{devlin2018bert,liu2019roberta,lan2019albert,reimers2019sentence,lewis2019bart}, Generative Pre-trained Transformers (GPT) models~\cite{radford2018improving,radford2019language,brown2020language}, etc. The building blocks of transformer networks are self-attention and feedforward layers that can be arranged in different ways to form the model, although the most common way is to stack up several blocks of these layers on top of each other.

The Vision Transformer (ViT) model introduced by Dosovitskiy \etal~\cite{dosovitskiy2020image} adopts a similar architecture as BERT for image classification with some modifications in the input layer. They conclude that ViT model, when trained on sufficiently large datasets, marginally outperforms ResNets in classifying Imagenet and CIFAR datasets. Touvron \etal~\cite{touvron2021training} improved the training efficiency of ViT using knowledge distillation ~\cite{hinton2015distilling} and devised Data-Efficient Image Transformer (DeiT). Their model was solely trained on a much smaller dataset, namely ImageNet1k (i.e., ILSVRC-2012 ImageNet dataset with 1k classes and 1.3M images)~\cite{dosovitskiy2020image,touvron2021training}. The key in their success is to utilize a simple yet effective knowledge distillation technique, that interestingly outperforms a vanilla distillation approach.
\section{Proposed Algorithm}\label{sec:3}
Our proposed PIRT model consists of three essential components–DeiT network, Proxy Anchor loss, and Soft Orthogonality (SO) constraint on proxies–each of which is described below.

\textbf{DeiT Network:}
We adopt the pre-trained, base version of DeiT~\cite{touvron2021training} as a vision transformer-based encoder in PIRT to extract rich contextual features from the input images. Its architecture consists of transformer blocks stacked on top of each other where each block is comprised of two sub-blocks: Multi-Head Self-attention (MHS) and Multi-Layer Perceptron (MLP). Furthermore, Layernorm (LN) and residual connections are applied before and after every sub-block, respectively. The MLP block itself consists of two layers with the Gaussian Error Linear Unit (GELU) activations.

Similar to ViT, the 2D input image with $(H, W)$ resolution and $C$ channels in DeiT is first split into N patches of $(P, P)$ resolution and $C$ channels, where $N = HW⁄P^{2}$ is the input sequence length. Then, the patch embeddings are created by flattening and mapping each patch to a fixed size $1D$ vector of size $D=C \times P \times P$. This process is similar to the tokenization in NLP, where each sentence or sequence is decomposed into the constituent tokens. To incorporate the position of patches in the initial image into the input signal to the network, a fixed positional embedding is added to the corresponding patch embedding vector.

The last step before sending the input to the model is to add two special, learnable tokens to the sequence of patch embeddings that are described below.

\emph{CLS token:} This token (aka., the Class token) is common to both DeiT and ViT, which is a learnable vector prepended to the sequence of patch embeddings.

\emph{ Distillation (DIST) token:} This token only exists in the DeiT architecture which is a trainable vector (just like CLS token) appended to the patch embeddings before the input layer and gets the supervisory signals from the teacher network during training.

The main difference between DeiT and ViT is the use of DIST token to implement the knowledge distillation through the teacher-student training strategy. Furthermore, DeiT uses a linear layer instead of MLP as the classification head on top of the CLS token during training. From the $N+2$ input tokens that DeiT processes, only the output states of CLS and DIST tokens are used for classification and other downstream tasks.

In~\cite{touvron2021training}, the teacher network for training DeiT is RegNetY-16GF~\cite{radosavovic2020designing}, which is trained separately with the same data and same data-augmentation as what used for DeiT. Furthermore, the following loss function called the hard-label distillation loss function~\cite{touvron2021training} is used for this teacher-student training framework:

\begin{equation}
	\mathcal{L}_{\text {global }}^{\text {hardDistill }}=\frac{1}{2} \mathcal{L}_{CE}\left(\psi\left(Z_{s}\right), y\right)+\frac{1}{2} \mathcal{L}_{C E}\left(\psi\left(Z_{s}\right), y_{t}\right),
	\label{e:hard_distill_loss}
\end{equation}
where $\mathcal{L}_{CE}$ is the cross-entropy loss, is the $\psi$ Softmax function, $y$ is the ground truth label, $y_{t}=argmax_{x}Z_{t}(c)$ is the hard decision of the teacher network, $Z_{s}$ and $Z_{t}$ are the logits of the student and the teacher networks, respectively.

To incorporate DeiT into PIRT, we remove the pre-trained classification linear heads from DeiT and replace them with a pooling layer to create a single vector representation of the input image. We have considered four different pooling methods in this study: 
\begin{itemize}
	\item \emph{Concat:} The concatenation of both CLS and DIST tokens,
	\item \emph{Mean:} This is the element-wise average of CLS and DIST tokens,
	\item \emph{CLS:} Using only the CLS token, and 
	\item \emph{DIST:} Using only the DIST token. 
\end{itemize}
Then, the pooled layer is linearly projected into a latent space to create an embedding vector with a desirable dimension.

\textbf{Proxy Anchor Loss:}
The Proxy Anchor loss~\cite{kim2020proxy} is a proxy-based loss that leads to concurrently learning two sets of vectors, that is, proxy vectors and embedding vectors of input data in a supervised learning framework. Here, each proxy represents a subset of data with a specific class label. For a given proxy of a particular class label, all inputs of the same class in the training batch are considered as positive examples and all other proxies and all other inputs are treated as negative proxies and examples, respectively. The Proxy Anchor loss is formulated as:
\begin{align}
	\ell_{\text {Proxy Anchor}} &= \frac{1}{\vert\mathcal{P}^{+}\vert} \sum_{p \in \mathcal{P}^+} \text{log}(1+\sum_{x \in \mathcal{X}_\mathcal{P}^{+}} \text{exp}(-\alpha (s(x,p)-\delta)))\nonumber\\
	& + \frac{1}{\vert\mathcal{P}\vert} \sum_{p \in \mathcal{P}} \text{log}(1+\sum_{x \in \mathcal{X}_\mathcal{P}^{-}}
	\text{exp}(\alpha (s(x,p)+\delta))),
	\label{e:proxy_loss}
\end{align}
where $s(\cdot,\cdot)$ represents the cosine similarity between two vectors, and $\alpha$, $\delta> 0$ denote the margin and scaling 
factors, respectively. $\mathcal{P}$ is the set of all proxies and $\mathcal{P}^{+}$ indicates the set of positive proxies in the batch of data. Furthermore, let $\mathcal{X}$ denotes the set of all embedding vectors of the inputs in the training batch. Then, $\mathcal{X}$ is divided into two separate sets for each proxy. $\mathcal{X}_\mathcal{P}^{+}$ is the set of the positive embedding vectors and $\mathcal{X}_\mathcal{P}^{-}$ is the set of the negative embedding vectors.

The Proxy Anchor loss function \eqref{e:proxy_loss} guides the network to learn the trainable parameters with a push-pull dynamic in the embedding space between the proxy vectors and the embedding vectors of the input images until an equilibrium reaches and the loss plateaus.

\textbf{Soft Orthogonality:}
One of the issues with the Proxy Anchor loss is that proxies are free to be as close as they can in the embedding space while they are representing different classes. This problem disrupts the push-pull dynamics. In other words, the proxies that are relatively close to each other have a hard time to pull the data points of the same class and push away other data points in the embedding space. To deal with this issue, we propose to impose the Soft Orthogonality (SO)~\cite{bansal2018can} as a regularization constraint on these proxies. The SO regularization requires the Gram matrix $P^TP$ of proxies to be close to unity using the Frobenius norm and is defined as:
\begin{equation}
	\left\|P^TP-I_r\right\|_{F}^{2},
	\label{eq:soft_orthogonality}
\end{equation}
in which $P \in \mathbb{R}^{d\times r}$ is the proxy matrix where each column represents a $d$-dimensional proxy vector, and $I_r$ is the identity matrix with the size $r$ such that $r$ is the number of proxies. Let us consider the column representation of $P$ as follows:
\begin{equation}\label{eq:soft_orthogonalityf112}
	P=\left[\begin{array}{cccc}
		p_{1} &
		p_{2}&
		\cdots&
		p_{r}
	\end{array}\right].
\end{equation}
Therefore, the Gram matrix $P^TP$ can be written as: 
\begin{equation}\label{eq:soft_orthogonalityf1}
	P^TP=\left[\begin{array}{cccc}
		p_{1}^Tp_{1} &
		p_{1}^Tp_{2}&
		\cdots&
		p_{1}^Tp_{r}\\ [1ex]
		p_{2}^Tp_{1} &
		p_{2}^Tp_{2}&
		\cdots&
		p_{2}^Tp_{r}\\[1ex]
		\vdots & \vdots & \ddots & \vdots \\[1ex]
		p_{r}^Tp_{1} &
		p_{r}^Tp_{2}&
		\cdots&
		p_{r}^Tp_{r}
	\end{array}\right].
\end{equation}
In an inner product space, like a Euclidean space of any dimension, the idea of orthogonality is closely linked to vector inner products. When the inner product of two vectors, such as $a$ and $b$, is zero, they are considered orthogonal. For instance, in Euclidean spaces, $\langle a,b\rangle=a^Tb=0$ implies the orthogonality, which geometrically implies that the vectors are at a 90-degree angle to each other and have no overlapping components. Two highly relevant concepts to orthogonality are similarity and redundancy. Similarity relates to the extent to which two or more objects are similar, and it can be used to detect trends or patterns in data. When two variables are similar, they can share common characteristics. Redundancy indicates the degree to which two or more variables overlap or duplicate each other. If two variables are redundant, they provide comparable information and may be used interchangeably. 

When dealing with vectors in inner product spaces such as $\mathbb{R}^d$, the cosine similarity $s(a,b)$ or the cosine of the angle $\theta_{a,b}$ between the two vectors $a$ and $b$ is defined as: 
\begin{equation}\label{cosinesim}
	s(a,b)=\cos(\theta_{a,b}) = \frac{a^Tb}{\Vert a\Vert_2 \Vert b\Vert_2},
\end{equation}
where $\Vert \cdot\Vert_2$ is the vector 2-norm. On this account, the similarity of two vectors can be assessed by measuring the degree to which their directions align closely. This can be quantified by computing their inner product or the angle between them. On the other hand, dissimilarity measures how dissimilar two vectors are and can be calculated as the inverse of their similarity. Consequently, if the inner product between two vectors is closer to zero, their angle will be closer to 90 degrees, indicating that their similarity and redundancy rate will be lower.

Now, putting the SO problem \eqref{eq:soft_orthogonality} and the relation \eqref{eq:soft_orthogonalityf1} together implies that 
\begin{equation}\label{eq:soft_orthogonalityf2}
	\left\|P^TP-I\right\|_{F}^{2}=\sum_{i,j=1, i\neq j}^r (p_i^Tp_j)^2+\sum_{i=1}^r (p_i^Tp_i-1)^2.
\end{equation}

When the problem \eqref{eq:soft_orthogonalityf2} is considered and it is assumed that it is being minimized with respect to $P$, the aim is to have $(p_i^Tp_j)^2$ equal to zero for every $i,j=1,\ldots,r$ such that $i\neq j$, and also to have $(p_i^Tp_i-1)^2$ equal to zero for every $i=1,\ldots,r$. Thus, by implementing the SO constraint within the Proxy Anchor loss, if the vectors $p_i$ and $p_j$ are trained to have an inner product that approaches zero, it implies that they have an orthogonal relationship. This indicates that they are dissimilar and non-redundant, with no shared information.

\textbf{PIRT's Loss Function:}
We formulate our loss function as a constrained optimization problem based on the SO constraint. It minimizes the following constrained Proxy Anchor loss to achieve the maximum distance among proxies representing dissimilar classes.
\begin{align}
	&\min_{P}\quad \mathcal{L}_{\text {PIRT }}=\ell_{\text {Proxy Anchor}},\nonumber \\
	&\,\mathrm{subject\,\,to}\,\,\,\,\,\, P^TP=I_r.
\end{align}
The Lagrangian function ${\mathcal{L}}$ of the constrained optimization of the above loss function can be written as follows:
\begin{equation}
	\begin{split}
		{\mathcal{L}}=\ell_{\text {Proxy Anchor }} + \lambda \left\| P^TP-I_r\right\|_{F}^{2},
	\end{split}
\end{equation}
where $\lambda>0$ is the penalty parameter.
\section{Experimental Results} \label{Experimental Results}
\textbf{Datasets:}
In this paper, Caltech-UCSD Birds-200-2011 (CUB-200-2011)~\cite{wah2011caltech}, Cars-196~\cite{krause20133d}, In-Shop Clothes Retrieval (In-Shop)~\cite{liu2016deepfashion}, and Stanford Online Products (SOP)~\cite{oh2016deep} datasets have been used to train and evaluate models. The CUB-200-2011 dataset is one of the most widely used datasets for deep metric learning, which contains $11,788$ images of $200$ different bird categories. The first $100$ categories with $5994$ images are considered for training and the remaining categories with $5794$ images are prepared for testing~\cite{hernandez2021attention}. The Cars-196 dataset contains $16,185$ car images of $196$ different model classes of which the first half are labeled for training and the second half for testing. The In-Shop benchmark contains images of clothing from different stores with various clothing types and poses. The training and test sets of this dataset are prepared differently. The training set includes $25,882$ images from $3,997$ different classes. The test set is split to query and gallery sets each of which has $3,985$, distinct classes. The former has $14,218$ images while the latter has $12,612$ images. The SOP dataset is the largest one in our experiments that has $120,053$ images of $22,634$ in different product classes. $11,316$ classes with $59,551$ images are assigned for training and the remaining classes with $60,502$ classes are considered for testing. All datasets introduced above that are analysed during the current study are available in public repositories which are listed in the Appendix.

\textbf{Evaluation Metrics:}
To evaluate the performance of our framework on the image retrieval task, we use Mean Average Precision at $R$ (MAP@R) and Precision@1 (P@1)~\cite{musgrave2020metric}. The MAP@R for a single query is defined as
\begin{equation}
	\mathrm{MAP} @ \mathrm{R}=\frac{1}{R} \sum_{i=1}^{R} P(i),
\end{equation}
where $R$ is the number of the nearest references to the query and $P(i)$ is the precision at $i$.

\textbf{Model Architecture:}
In our experiments, we utilized the base version of the pre-trained DeiT model, which is referred to as DeiT-B in~\cite{touvron2021training}, as the encoder backbone. In addition to our proposed PIRT model, we examined two different variations of PIRT to compare their performance with other deep learning-based image retrieval models. We investigated the sensitivity of PIRT to the backbone by substituting DeiT with Resnet50, and we also evaluated the performance of our proposed method against the robust MS loss~\cite{wang2019multi}. Along with our proposed PIRT model, there are three other models used as baselines to perform a sensitivity analysis with respect to different backbones and loss functions, which are summarized in Table~\ref{Table1}.

\begin{table}[!htbp]	 
		\centering
		\caption{Sensitivity analysis of our proposed method PIRT with respect to different backbones, loss functions, and soft orthogonality.}
		\resizebox{\linewidth}{!}{
		\begin{tabular}{ccccp{10cm}}
			\hline
			\multicolumn{1}{|c|}{\textbf{Model Name}} & \multicolumn{1}{c|}{\textbf{Neural Network Backbone}} & \multicolumn{1}{c|}{\textbf{Loss Function}} & \multicolumn{1}{c|}{\textbf{Soft Orthogonality}} \\ \hline
			\\[-3mm]\hline
			\multicolumn{1}{|c|}{PIRT} & \multicolumn{1}{c|}{DeiT-B} & \multicolumn{1}{c|}{Proxy Anchor} & \multicolumn{1}{c|}{Yes} \\
			\multicolumn{1}{|c|}{DeiT-PA} & \multicolumn{1}{c|}{DeiT-B} & \multicolumn{1}{c|}{Proxy Anchor} & \multicolumn{1}{c|}{No} \\
			\multicolumn{1}{|c|}{DeiT-MS} & \multicolumn{1}{c|}{DeiT-B} & \multicolumn{1}{c|}{MS} & \multicolumn{1}{c|}{N/A} \\
			\multicolumn{1}{|c|}{Resnet50-PA} & \multicolumn{1}{c|}{Resnet50} & \multicolumn{1}{c|}{Proxy Anchor} & \multicolumn{1}{c|}{No} \\
			\multicolumn{1}{|c|}{Resnet50-MS} & \multicolumn{1}{c|}{Resnet50} & \multicolumn{1}{c|}{MS} & \multicolumn{1}{c|}{N/A} \\ \hline
		\end{tabular}
			}
	\label{Table1}
\end{table}

\textbf{Training Configurations:}
We employed the AdamW optimizer~\cite{loshchilov2017decoupled} with a weight decay of $10^{-4}$ and an initial learning rate of $10^{-4}$ (unless otherwise specified) for all models. The StepLR learning rate scheduler~\cite{paszke2019pytorch} was utilized to decay the learning rate by a factor of $0.5$ every $5$ epochs. The number of training epochs was set to $30$ for the CUB-200-2011 and Cars-196 datasets, and $60$ for the In-Shop and SOP datasets. It is worth noting that we allocated the first 5 epochs of all training tasks for warm-up. Furthermore, to expedite convergence~\cite{kim2020proxy}, we increased the learning rate for the proxies by 100 times, in both the PIRT and Resnet50-PA networks.

\textbf{Loss Function Settings:}
In order to prevent the creation of redundant proxies in the proxy-based approaches, such as PIRT and Resnet50-PA, we generated a single proxy vector for each class present in the training data. The initial values of these vectors were drawn from a normal distribution in such a way that the proxies were uniformly distributed on the unit hypersphere. Additionally, we set $\alpha$ and $\delta$ in Eq.\ref{e:proxy_loss} to $32$ and $0.1$, respectively, as recommended by previous work\cite{kim2020proxy}. For the MS-based methods, such as DeiT-MS and Resnet50-MS, we used the same hyperparameters for the MS model as those suggested in~\cite{wang2019multi}.

\textbf{Input Settings:}
To address overfitting and improve the generalization of the trained encoders, we utilized various data augmentation methods on the input images. Specifically, during training, we applied random cropping with resizing to $224\times 224$ pixels and random horizontal flipping to the images. During inference, we first resized the input images to $256\times 256$ pixels, and then performed center-cropping with resizing to $224\times 224$ pixels.

\textbf{Quantitative Results:}
Our primary aim is to showcase the superior performance of the PIRT framework for image retrieval relative to other cutting-edge techniques. To ensure equitable comparisons, we kept the image size constant at $224 \times 224$ for all datasets. Furthermore, we assessed the efficacy of our framework using an embedding dimension of $512$.
We commence with quantitative comparisons by presenting the P@1 values in Table~\ref{Table2} for all four datasets. The first column indicates the method's name, while the second to last columns reveal the P@1 performance for Cars196, CUB, In-Shop, and SOP datasets, respectively. The methods are categorized based on their backbones. For instance, methods that utilize Resnet50 as their backbone are separated from other methods with different backbones. Please note that the P@1 values for Inception backbone with Proxy Anchor loss (Inception-PA) and MS loss (Inception-MS) are directly extracted from~\cite{kim2020proxy}. 
\begin{table}[!htbp]
	\centering
	\caption{P@1 on the CUB-200-2011, Cars196, In-shop, and SOP. The best performance is bolded. The subscripts are showing the
		$\lambda$ coefficient value in the loss function.}
	\resizebox{\linewidth}{!}{	
	\begin{tabular}{|l|c c c c|}
		\hline
		\textbf{Model} & \textbf{CUB-200-2011} & \textbf{Cars196} & \textbf{In-Shop} & \textbf{SOP} \\ \hline
		Resnet50-PA & 69.5 & 86.1 & 84.6 & 76.2 \\
		Resnet50-MS & 67.4 & 82.1 & 66.1 & 70.4 \\
		Resnet50-ProxyNCA++ \cite{wern2020proxynca++} & 69.0 & 86.5 & 90.4 & 80.7 \\ \hline
		Inception-PA \cite{kim2020proxy} & 68.4 & 86.1 & 91.5 & 79.1 \\
		Inception-MS \cite{kim2020proxy} & 65.7 & 84.1 & 89.7 & 78.2 \\
		Inception-HTL \cite{ge2018deep} & 57.1 & 81.4 & 80.9 & 74.8 \\
		Inception-RLL-H \cite{wang2019ranked} & 49.2 & 73.2 & N/A & 76.1 \\
		Inception-SoftTriple \cite{qian2019softtriple} & 49.8 & 64.7 & N/A & 78.3 \\ \hline
		GoogleNet-HDC \cite{yuan2017hard} & 53.6 & 73.7 & N/A & 69.5 \\
		GoogleNet-A-BIER \cite{opitz2018deep} & 57.5 & 82.0 & 83.1 & 74.2 \\
		GoogleNet-ABE \cite{kim2018attention} & 60.6 & 85.2 & 87.3 & 76.3 \\ \hline
		DeiT-PA  & 79.9 & 90.7 & 91.0 & 84.4 \\
		DeiT-MS  & 79.6 & 89.6 & 82.0 & 79.1 \\
		PIRT\textsubscript{0.1}  & 79.7 & \textbf{91.3} & 91.5 & 82.5 \\
		PIRT\textsubscript{0.01}  & 79.8 & 91.0 & 91.8 & 84.6 \\
		PIRT\textsubscript{0.001}  & \textbf{80.0} & 90.8 & \textbf{92.0} & \textbf{84.8} \\ \hline
	\end{tabular}
}
	\label{Table2}
\end{table}

Table~\ref{Table2} demonstrates that our model, PIRT, achieves state-of-the-art performance by outperforming all other models in terms of P@1 values across all datasets. Our method substantially outperforms other non-transformer approaches (rows 1 to 11 in Table~\ref{Table2}) with a large margin, while only slightly outperforming transformer-based approaches. This observation suggests that transformers can capture richer global representations of the input compared to off-the-shelf CNNs.
Also, Table~\ref{Table2} suggests that our model, PIRT, significantly outperforms Inception-PA in terms of P@1 with a large margin ($11.6\%$ in CUB-200-2011, $5.2\%$ in Cars196, $0.5\%$ in In-Shop, and $5.7\%$ in SOP). Similarly, our method outperforms Inception-MS across all datasets by a large margin.
It is worth noting that the DeiT-PA model performs better than other approaches on all datasets, except for In-Shop, where Inception-PA achieves a slightly higher P@1 value by $0.5\%$. However, when we incorporate soft orthogonality (SO) into DeiT-PA, creating the PIRT model, it outperforms Inception-PA for the In-Shop dataset regarding P@1. Additionally, models based on Proxy Anchor loss, whether with or without a vision transformer as the encoder backbone, demonstrate higher accuracy than models based on MS loss across all datasets, except for the In-Shop dataset. For the In-Shop dataset, the P@1 value of Inception-MS is higher than that of Resnet50-PA, Resnet50-MS, and DeiT-MS. However, the P@1 value of Inception-MS for the In-Shop dataset is lower than that of DeiT-PA and PIRT.

\begin{table}[!htbp]
	\caption{Comparisons of evaluation metrics between PIRT and DeiT-PA models, while applying four different pooling methods to the special tokens at the top layer of each model.}
\resizebox{\linewidth}{!}{
\begin{tabular}{|c|c|cc|cc|cc|}
			\cline{3-8}
			\multicolumn{2}{c|}{} & \multicolumn{2}{c|}{\textbf{CUB-200-2011}} & \multicolumn{2}{c|}{\textbf{Cars-196}} & \multicolumn{2}{c|}{\textbf{In-Shop}} \\ \hline
			\textbf{Model Framework} & \textbf{Pooling Method} & P@1 & MAP@R & P@1 & MAP@R & P@1 & MAP@R \\ \hline
		PIRT & Concat & 80.0 & 38.7 & 90.8 & 34.3 & 92.0 & 65.6 \\
		(reg\_coeff=0.001)	& Mean & 79.6 & 38.0 & 90.7 & 34.4 & 92.0 & 65.6 \\
			& CLS & 78.6 & 37.3 & 90.3 & 33.8 & 91.8 & 65.2 \\
			& DIST & 79.7 & 38.5 & 91.3 & 35.1 & 92.0 & 65.8 \\ \hline
			DeiT-PA & Concat & 79.9 & 38.8 & 90.7 & 34.5 & 91.0 & 64.7 \\
			& Mean & 79.5 & 38.2 & 90.6 & 34.5 & 91.2 & 64.8 \\
			& CLS & 78.6 & 37.4 & 90.1 & 33.9 & 91.1 & 64.4 \\
			& DIST & 79.8 & 38.8 & 91.3 & 35.1 & 91.2 & 65.0 \\ \hline
		\end{tabular}}
	\label{Table3}
\end{table}

Table~\ref{Table3} displays the performance of our model in comparison to baseline models based on MAP@R. Our model outperforms all baseline models in terms of MAP@R for all datasets except for CUB-200-2011. Additionally, we observed that PIRT outperforms DeiT-PA in terms of MAP@R for Cars-196, In-Shop, and SOP datasets, with MAP@R values of PIRT being $0.3\%$, $0.9\%$, and $0.8\%$ larger, respectively. However, for the CUB-200-2011 dataset, MAP@R of PIRT is slightly smaller than that of DeiT-PA by $0.1\%$. The results suggest that Soft Orthogonality is effective in improving image retrieval performance since both PIRT and DeiT-PA use the same encoder (DeiT). The only difference between the two is our constrained loss function.

It is important to note that the full capabilities of our proposed loss function are demonstrated in the CUB-200-2011 and Cars-196 datasets, where the constraint in our loss is never low-ranked due to the significantly smaller number of proxies compared to the embedding dimensions. However, in larger datasets such as In-Shop and SOP, our constraint in the loss function becomes low rank as the number of proxies exceeds the embedding dimensions.

 \textbf{Pooling Methods:}
In this analysis, we investigate the influence of different pooling methods on the retrieval performance of PIRT (with $\lambda=0.001$) and DeiT-PA models on three datasets. Table~\ref{Table3} presents the accuracies of both models with four different pooling methods. The results indicate that the Concat pooling method achieves better retrieval performance for both models applied to the CUB-200-2011 dataset, while the DIST pooling method outperforms the other pooling methods for the Cars-196 and In-Shop datasets. On the other hand, the CLS pooling method yields the worst performance in all three datasets, especially in the CUB-200-2011 and Cars-196 datasets for both PIRT and DeiT-PA models. Specifically, PIRT with CLS pooling has P@1 and MAP@R values $1.4\%$ smaller than those of PIRT with Concat pooling in the CUB-200-2011 dataset. Furthermore, PIRT with CLS pooling for the Cars-196 dataset has P@1 and MAP@R values smaller than those of PIRT with DIST pooling by $1\%$ and $1.3\%$, respectively. These observations suggest that the DIST token contains useful information learned from the teacher network during the pre-training of the DeiT model using the hard-label distillation loss in Eq.~\ref{e:hard_distill_loss}.

\begin{figure}[!htbp]
    \centering
    \raisebox{-\height}{\includegraphics[width=0.95\linewidth]{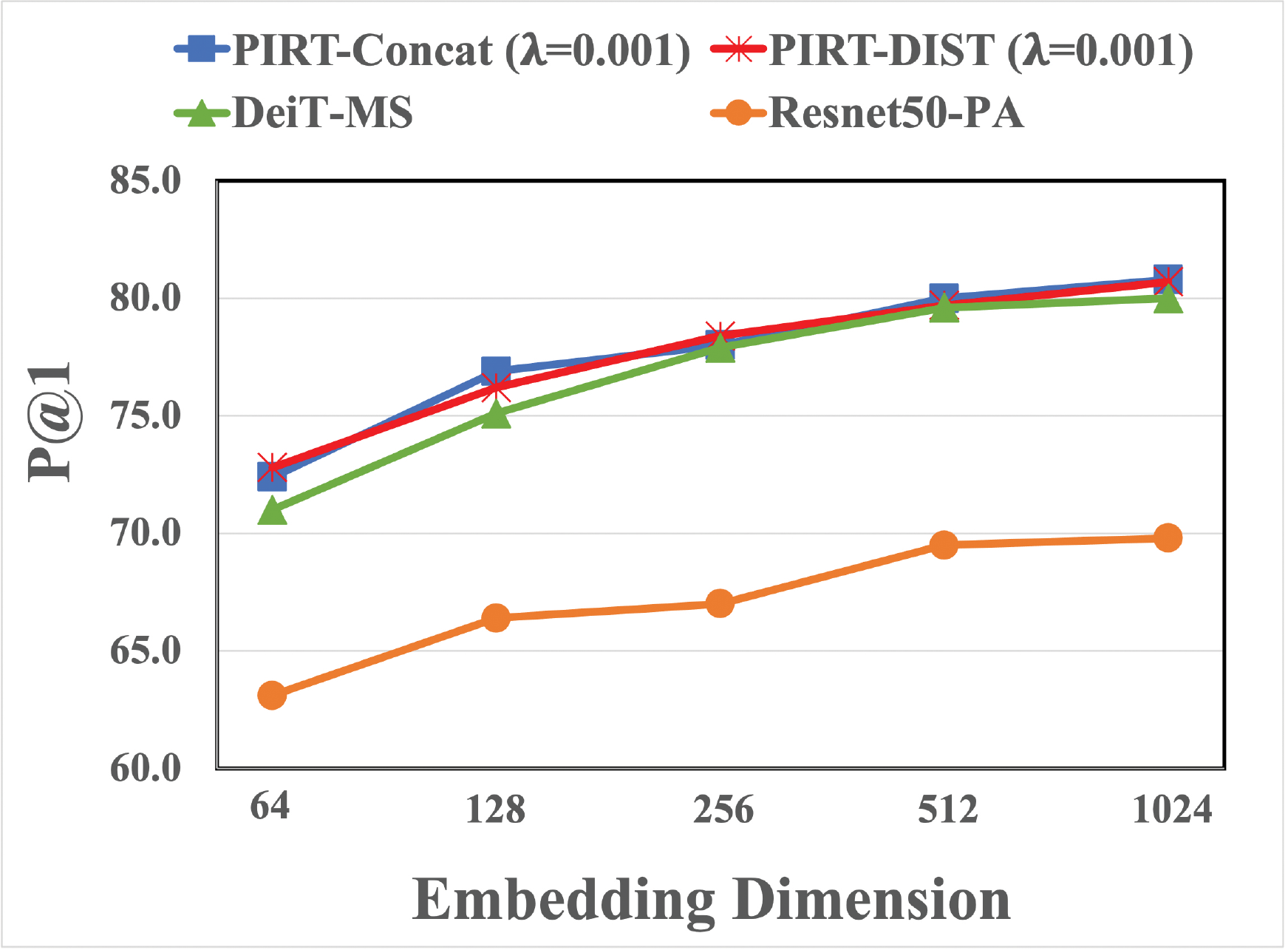}}
    \hfill
    \raisebox{-\height}{\includegraphics[width=0.95\linewidth]{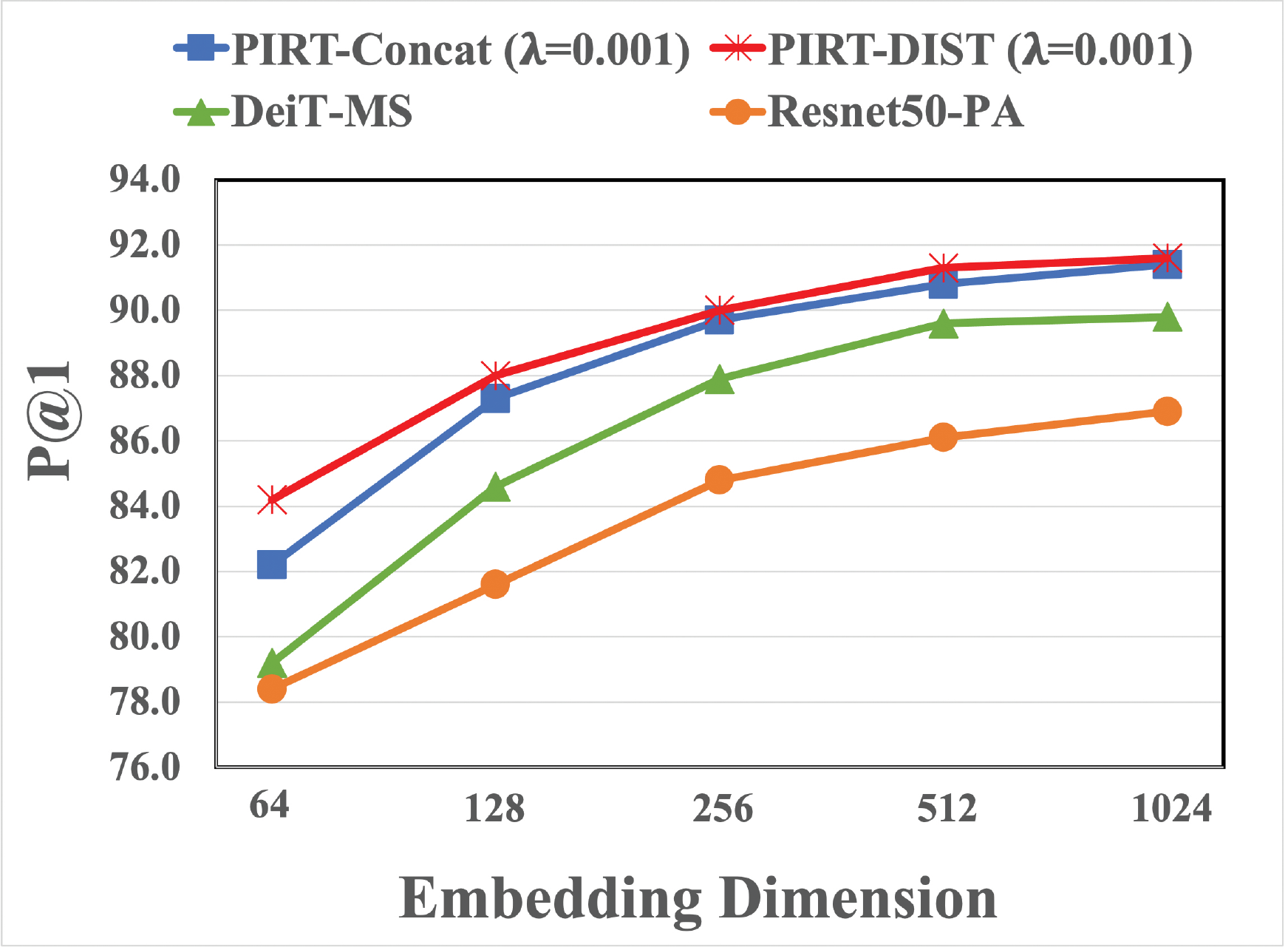}}
 	\caption{The effects of embedding dimension on the accuracy of models for a) CUB-200-2011 dataset and b) Cars-196 dataset. PIRT-Concat and PIRT-DIST are two different versions of PIRT that use Concat and DIST pooling methods, respectively.}
 	\label{fig1_fig2}
\end{figure}

 \textbf{Embedding Dimension:}
The impact of embedding vector dimension on image retrieval accuracy is critical, affecting the trade-off between accuracy and speed. To investigate this, we varied the embedding size of PIRT models from 64 to 1024 and used Concat and DIST pooling methods with $\lambda=0.001$, inspired by~\cite{kim2020proxy}. Fig~\ref{fig1_fig2} displays the results for CUB-200-2011 and Cars-196 datasets, with the accuracies of DeiT-MS and Resnet50-PA models also shown.
%

In both datasets, PIRT models achieve significantly higher accuracies than Resnet50-PA for all embedding dimension values. In the Cars-196 dataset, PIRT models also outperform DeiT-MS for all embedding dimensions. Notably, the performance of PIRT models on the Cars-196 dataset, particularly with DIST pooling, declines at a slower rate below 128 embedding dimensions compared to DeiT-MS and Resnet50-PA. Additionally, while PIRT-DIST outperforms PIRT-Concat in the Cars-196 dataset, their difference in retrieval accuracy becomes less significant for embedding dimensions above 512. For the CUB-200-2011 dataset, the accuracies of DeiT-MS are closer to those of PIRT models compared to Resent50-PA, yet DeiT-MS under-performs compared to PIRT models, particularly for embedding dimensions below 128. Nevertheless, higher embedding dimensions lead to improved image retrieval accuracies in both datasets.

\begin{figure}[!htbp]
    \centering
    \raisebox{-\height}{\includegraphics[width=0.95\linewidth]{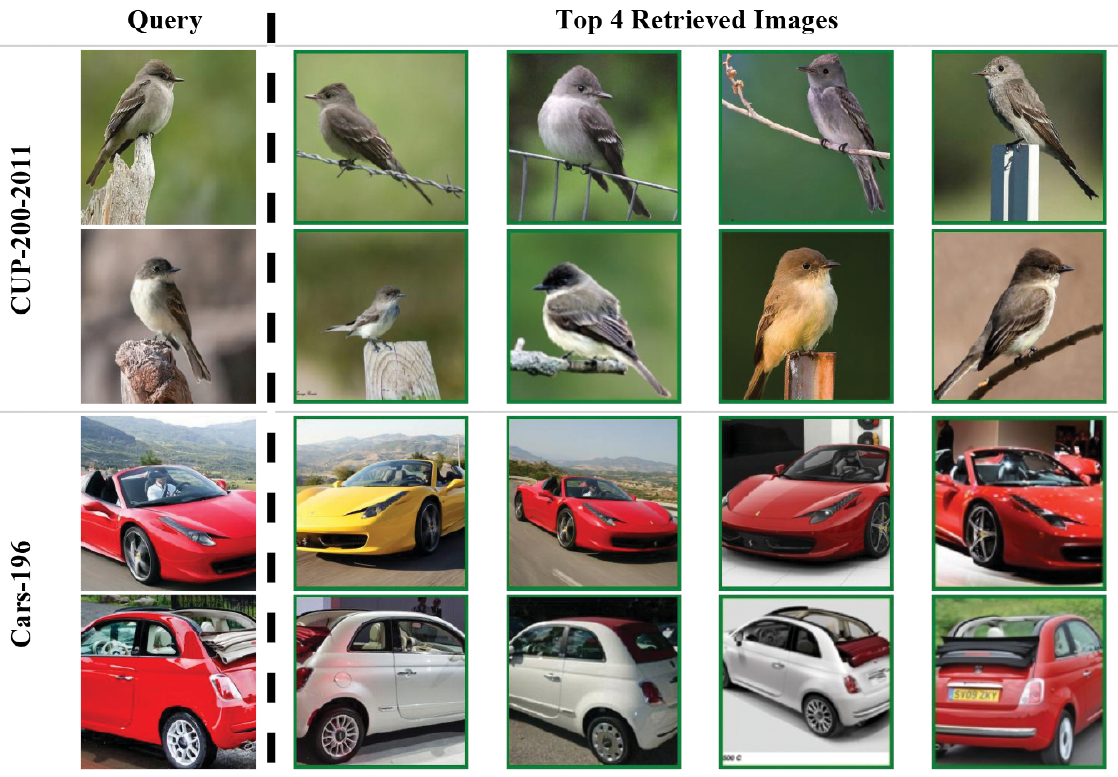}}
    \hfill
    \raisebox{-\height}{\includegraphics[width=0.95\linewidth]{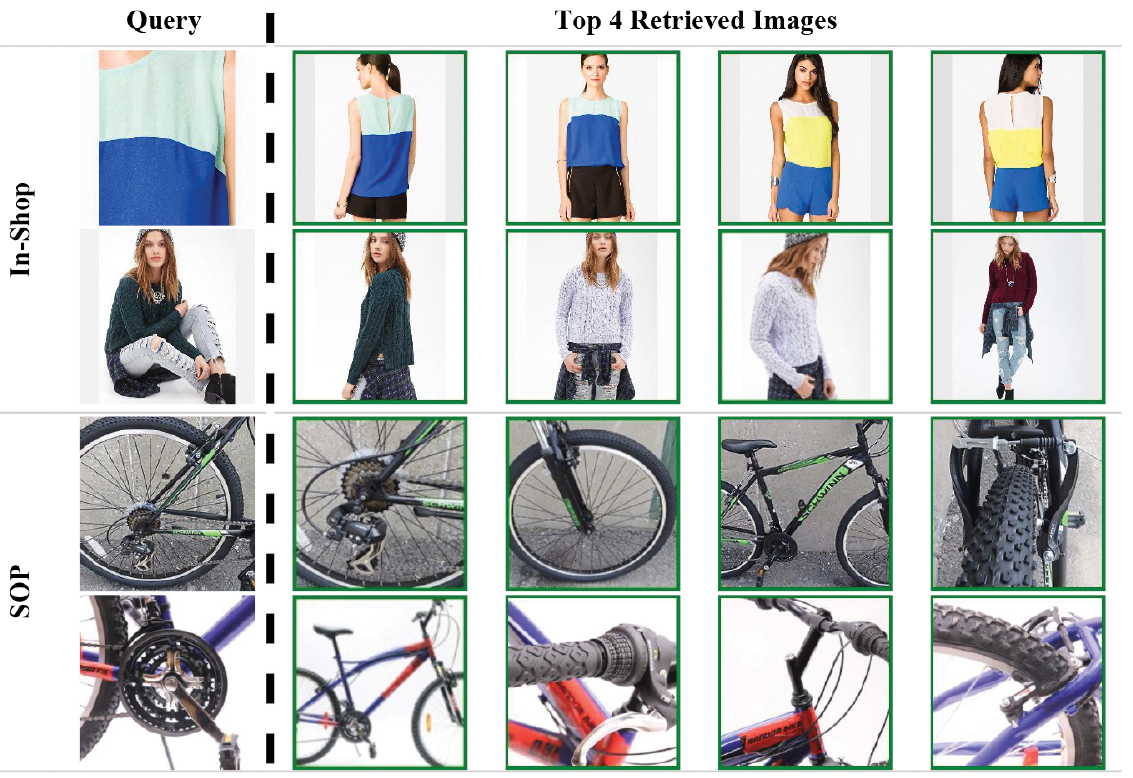}}
      	\caption{Top 4 images retrieved from four different datasets by PIRT model for two different query images. The green border indicates the true retrieval.}
  	\label{fig5}
\end{figure}

\begin{figure*}[!htbp]
    \centering
    \begin{subfigure}[b]{0.45\linewidth}
        \centering
        \includegraphics[width=\linewidth]{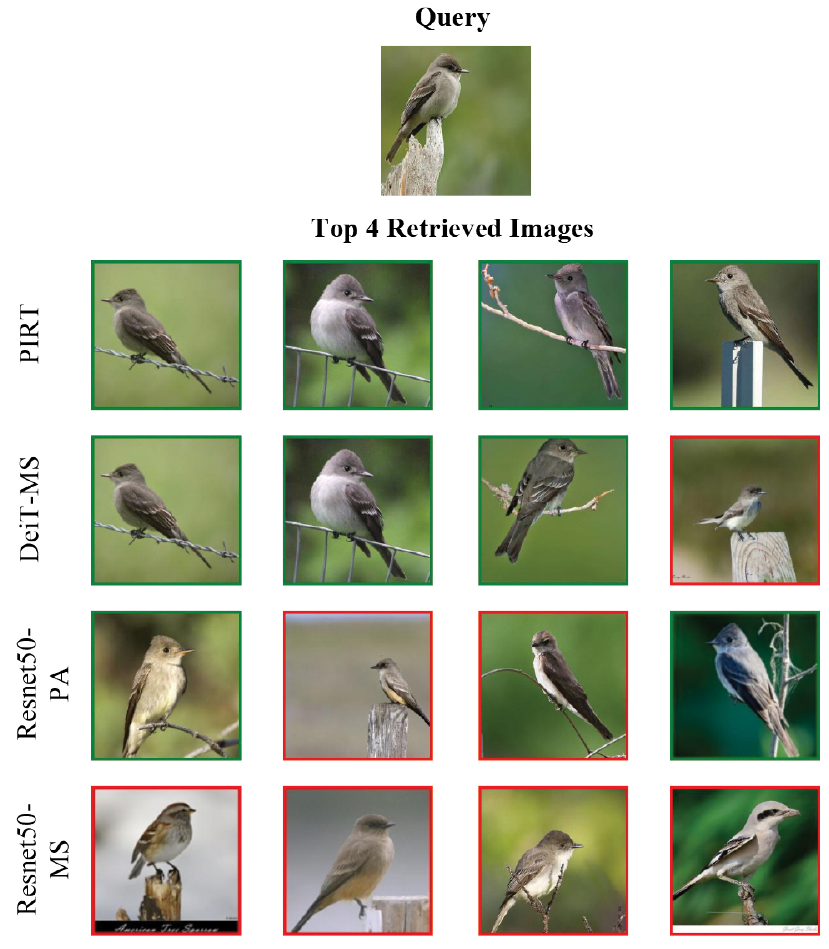}
        \caption{}
        \label{fig:sub-a}
    \end{subfigure}
    \hfill
    \begin{subfigure}[b]{0.45\linewidth}
        \centering
        \includegraphics[width=\linewidth]{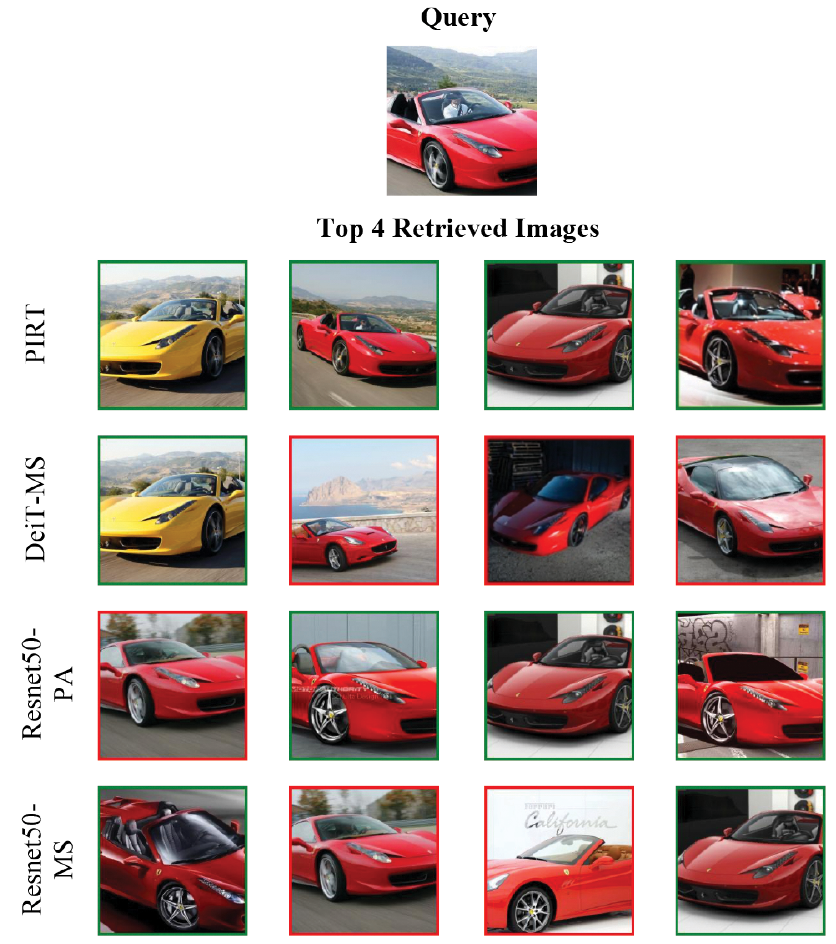}
        \caption{}
        \label{fig:sub-b}
    \end{subfigure}
    \\
    \begin{subfigure}[b]{0.45\linewidth}
        \centering
        \includegraphics[width=\linewidth]{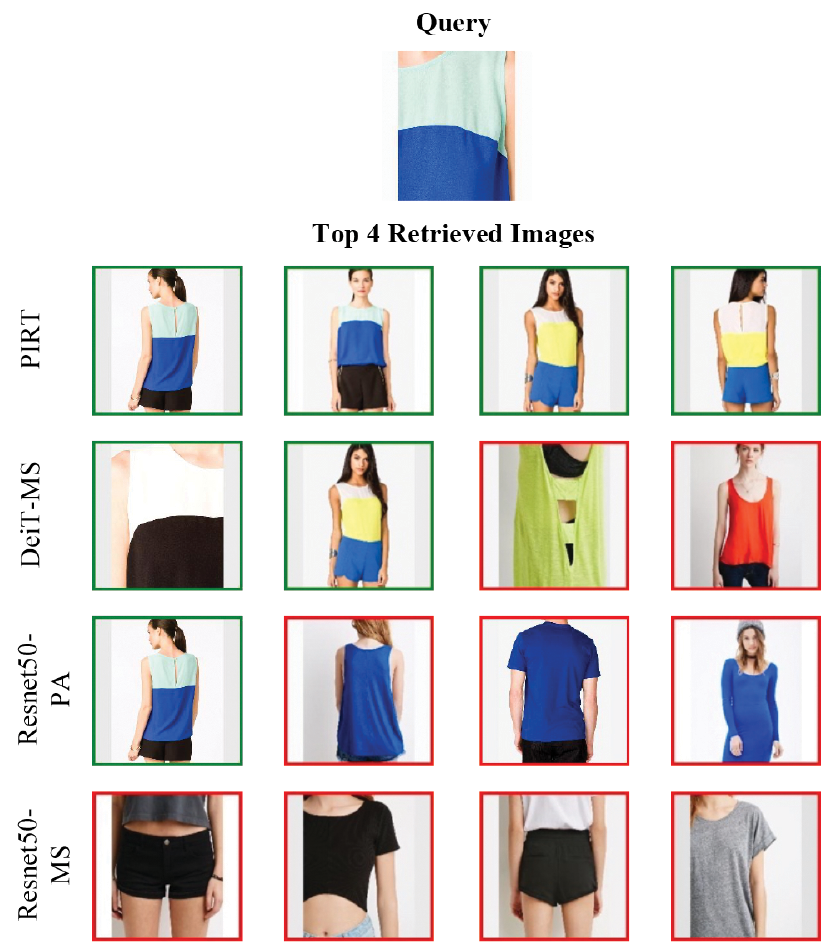}
        \caption{}
        \label{fig:sub-c}
    \end{subfigure}
    \hfill
    \begin{subfigure}[b]{0.45\linewidth}
        \centering
        \includegraphics[width=\linewidth]{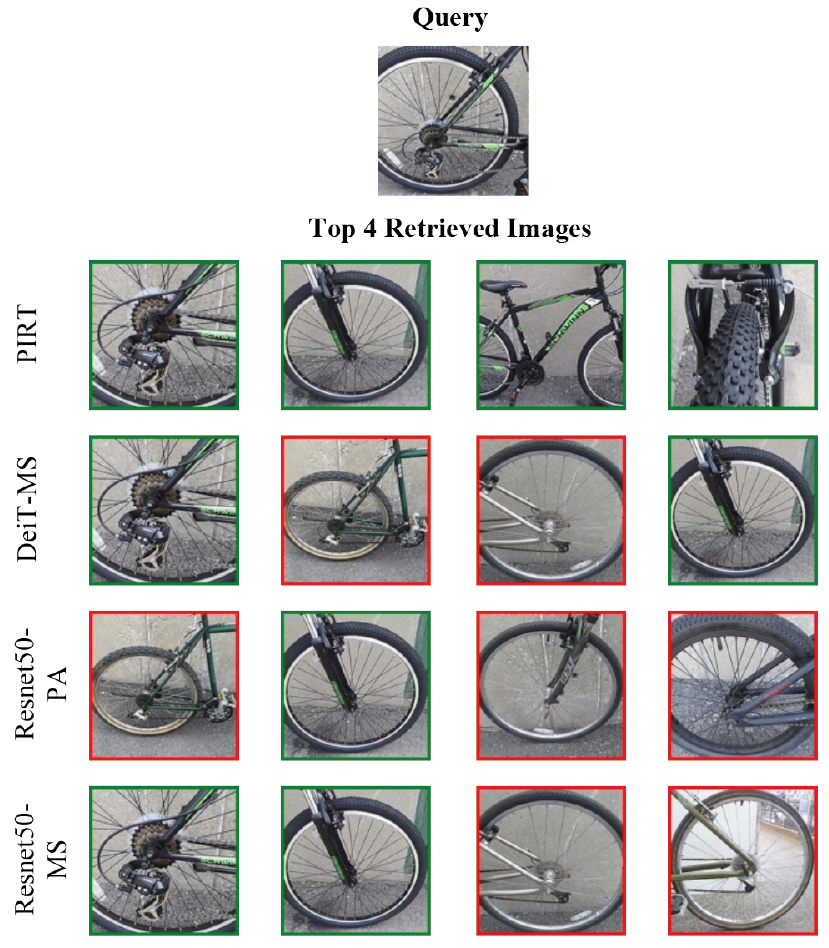}
        \caption{}
        \label{fig:sub-d}
    \end{subfigure}
    \caption{Top 4 images retrieved by four different models for a given query image from four different datasets: (a) CUB-200-2011, (b) Cars-196, (c) In-Shop, and (d) SOP. The green border indicates the true retrieval.}
    \label{fig6}
\end{figure*}

 \textbf{Qualitative Results:}
Here, we aim to visually investigating the performance of the PIRT model in retrieving images from the four datasets for a given image query. Fig~\ref{fig5} illustrates the top 4 images (ranked based on their MAP@R) retrieved by PIRT model for two different queries from CUB-200-2011, Cars-196, In-Shop, and SOP datasets. To compare the image retrieval performance of other models with PIRT, the first query for each dataset in Fig~\ref{fig5} was passed through PIRT, DeiT-MS, Resnet50-PA, and Resnet50-MS. The top 4 images (ranked based on their MAP@R) retrieved by each model are shown in Fig~\ref{fig6}. 

A striking observation in Fig~\ref{fig5} is the large intra-class variations, which makes the task of image retrieval challenging. For instance, the CUB-200-2011 dataset displays different poses and background clutter and the images in Cars-196 dataset contain objects with various colors and shapes. The view-point changes in In-Shop and SOP datasets introduce even more challenges to the image retrieval task. Nevertheless, the PIRT model demonstrates an outstanding performance in retrieving images through projecting the images into an embedding space where intrigue subtleties in the images can be captured. 

Let us examine the first query in the In-Shop dataset in Fig~\ref{fig5}, which contains a two-tone women dress in light blue-dark blue texture. The first top 2 retrieved images for this query have similar textures and colors but with different poses. The last top 2 retrieved images have also the similar textures to the query but with different two-tone colors in white and yellow. Furthermore, it is worth comparing the results in Fig~\ref{fig6}(c) pertaining to the image retrieval of the mentioned query by four different models of PIRT, DeiT-MS, Resnet50-PA, Resnet50-MS. The first retrieved image of PIRT and Resnet50-PA, both of which use Proxy Anchor in their loss function, are the same. However, the other three retrieved images of Resnet50-PA are not relevant to the given query at all. The first two retrieved image of DeiT-MS are relevant to the query, although they have different colors than that of the query image. But the other two images are not retrieved correctly by this model. Although we can see two different colors in these two images, they are not part of a single cloth as opposed to the two-tone color in the query image. All images retrieved by Resnet50-MS are non-relevant to the query image.

The ability of the PIRT model in capturing the context in the images is also evident in the retrieved images of SOP dataset in Fig~\ref{fig5}, where both queries contain the visual details of a small section of two different bikes. Both query images contain a combination of contrasting colors that interestingly exist in the retrieved images as well. For instance, the second query of SOP includes red, purple, and black colors. All the corresponding top 4 retrieved images also contain this combination of colors. Fig~\ref{fig6}(d) shows that DeiT-PA, Resnet50-PA, and Resnet50-MS models miss these color combinations in some of the retrieved images from SOP dataset for the same query image.

 \section{Conclusion}\label{concsec}
In this paper, we have proposed PIRT as a novel proxy-based image retrieval framework. Our proposed method leverages a vision transformer as well as a proxy-based loss function regularized by soft orthogonality. The vision transformer extracts enriched contextual features from the input images thanks to its powerful self-attention mechanism. Then, the extracted features are passed along to the projection layer. While the pre-trained vision transformer is maintained frozen, the constrained Proxy Anchor loss helps the projection layer learn a substantially well embedding vector. We extensively evaluated our proposed method on four publicly available image retrieval datasets. Our results confirm that our proposed method is outperforming state-of-the-art methods often by a large margin. For future work, although our proposed Soft Orthogonality constraint on Proxy Anchor loss showed some improvements, an exploration of other ways of constraining proxies could be considered.

 \section{Ethics declarations}
No conflict of interest exists. Dr. Farshad Saberi-Movahed contributed to this article in his personal capacity. The views expressed are his own and do not necessarily represent the views of NVIDIA.


{\small
 
}

\section{Appendix}
This Appendix presents the public repositories of the analyzed data in our work as well as some additional experimental results that were excluded from the main paper. The public repositories of the datasets, all of which were introduced in Section ~\ref{Experimental Results}, are listed below.
\begin{itemize}
  \item CUB-200-2011: \url{http://www.vision.caltech.edu/datasets/cub_200_2011/}
  \item Cars-196: \url{https://ai.stanford.edu/~jkrause/cars/car_dataset.html}
  \item In-Shop: \url{http://mmlab.ie.cuhk.edu.hk/projects/DeepFashion.html}
  \item SOP: \url{https://cvgl.stanford.edu/projects/lifted_struct/}
\end{itemize}

\end{document}